\documentclass[a4paper,11pt,twocolumn,twoside]{article}

\usepackage[utf8]{inputenc}
\usepackage[T1]{fontenc}
\usepackage[english]{babel}

\usepackage{graphicx}
\usepackage[table,xcdraw]{xcolor}
\usepackage{amsmath, amsfonts}
\usepackage{booktabs}
\usepackage{tikz}
\usepackage{pgfplots}
\usepackage{caption}
\usepackage{listings}
\usepackage{subcaption}
\usepackage{hyperref}
\definecolor{codegreen}{rgb}{0,0.6,0}
\definecolor{codegray}{rgb}{0.5,0.5,0.5}
\definecolor{codepurple}{rgb}{0.58,0,0.82}
\definecolor{backcolour}{rgb}{0.95,0.95,0.92}
 
\lstdefinestyle{codigo}{
    backgroundcolor=\color{white},
    commentstyle=\color{codegreen},
    keywordstyle=\color{blue},
    numberstyle=\tiny\color{codegray},
    stringstyle=\color{codepurple},
    basicstyle=\ttfamily\footnotesize,
    breakatwhitespace=true,   
    breaklines=true,                 
    captionpos=b,  
    columns=fullflexible,
    keepspaces=true,                 
    numbersep=5pt,                  
    showspaces=false,
    showstringspaces=false,
    showtabs=false,                  
    tabsize=2
}
\lstdefinestyle{pseudo}{
    backgroundcolor=\color{backcolour},
    basicstyle=\ttfamily\footnotesize,
    numberstyle=\tiny\color{codegray},
    breakatwhitespace=false,         
    breaklines=true,                 
    captionpos=b,                    
    keepspaces=true,                 
    numbers=left,                    
    numbersep=5pt,                  
    showspaces=false,                
    showstringspaces=false,
    showtabs=false,                  
    tabsize=2
}

\definecolor{codegreen}{rgb}{0,0.6,0}
\definecolor{codegray}{rgb}{0.5,0.5,0.5}
\definecolor{codepurple}{rgb}{0.58,0,0.82}
\definecolor{backcolour}{rgb}{0.94,0.97,1.0}
\lstdefinestyle{xmlstyle}{
    backgroundcolor=\color{backcolour},   
    commentstyle=\color{codegreen},
    keywordstyle=\color{magenta},
    numberstyle=\tiny\color{codegray},
    stringstyle=\color{codepurple},
    basicstyle=\ttfamily\scriptsize,
    breakatwhitespace=false,         
    breaklines=true,                 
    captionpos=b,                    
    keepspaces=true,                 
    numbers=left,                    
    numbersep=5pt,                  
    showspaces=false,                
    showstringspaces=false,
    showtabs=false,                  
    tabsize=2,
    upquote=true,
    language=xml
}
\hypersetup{
  colorlinks,
  citecolor=black,
  filecolor=black,
  linkcolor=black,
  urlcolor=black
}

\title{Fine-tuning of Large Language Models for Constituency Parsing Using a Sequence to Sequence Approach}

\author{
\textbf{Francisco José Cortés Delgado}$^{1}$,
\textbf{Eduardo Martínez Graciá}$^{1}$,
\textbf{Rafael Valencia García}$^{1}$\\[0.5em]
$^{1}$University of Murcia, Spain\\
\texttt{franciscojose.cortesd@um.es, edumart@um.es, valencia@um.es}
}

\date{}

\begin{document}
\maketitle

\begin{abstract}
Recent advances in natural language processing using large neural models enable investigating a new syntactic approach to phrase-structure analysis based on machine learning. This work proposes fine-tuning large language models for phrase-structure analysis by translating an input sequence (the sentence to be analyzed) into an output sequence (its phrase-structure analysis). The ultimate goal of this technique is to expand the functionalities of the \textit{MiSintaxis} tool, designed for teaching Spanish syntax. Models available in Hugging Face have been fine-tuned on training data generated from the AnCora-ES corpus, and the results have been compared using the $F_1$ metric. The results indicate high precision in syntactic phrase-structure analysis while highlighting this methodology's potential.
\end{abstract}


\section{Introduction}
Traditionally, constituency parsing has employed techniques based on the Cocke–Younger–Kasami (CYK) algorithm. However, the complexity and ambiguity inherent to natural languages have posed significant challenges to this approach. This work proposes a new method based on the automatic analysis of the grammatical structure of Spanish sentences using large language models (LLMs), such as Bloom or GPT-2, available on Hugging Face. These models can be fine-tuned to perform the constituency parsing task through an approach known as \emph{sequence-to-sequence translation}. With the fine-tuned models, it becomes possible to integrate an automatic syntactic parser into educational tools for teaching Spanish syntax, such as \cite{misintaxis}, an application designed for pre-university students that currently has thousands of users worldwide.

Section \ref{sec:relacionado} summarizes the most relevant references on which this work is based. Section \ref{sec:desarrollo} describes the preparation and evaluation of the fine-tuned models. Finally, Section \ref{sec:conclusiones} presents the conclusions and outlines future work.

\section{Related Work}\label{sec:relacionado}
The most important traditional tools for constituency parsing have been context-free grammars in Chomsky Normal Form and algorithms based on the Cocke–Younger–Kasami (CYK) dynamic programming method, which allow the binary structure of the syntactic tree of a given sentence to be established. The main difficulty of this approach lies in constructing a grammar expressive enough to describe all the complex syntactic phenomena of a natural language.

In the past decade, the use of deep neural networks has grown significantly across many natural language processing (NLP) applications, particularly since the development of the \emph{self-attention} mechanism, which gave rise to modern large language models. This mechanism, replicating human cognitive attention, was introduced by \cite{NIPS2017_3f5ee243}, surpassing the effectiveness of previous techniques used in \emph{Long Short-Term Memory} (LSTM) networks. Through attention, present in the neural architecture known as the \emph{transformer}, a part of the network can determine which portions of the previous context are most relevant to continue generating the output during the inference process.

Attention mechanisms are not only useful for morphological word generation but also appear to learn the syntactic categories of language, as suggested by \cite{DBLP:journals/corr/abs-1911-03875}. For this reason, researchers have begun to apply them to constituency parsing. \cite{DBLP:journals/corr/VinyalsKKPSH14} proposed that constituency parsing can be approached similarly to language translation, using a \emph{sequence-to-sequence} framework. Given an input sequence (the unparsed sentence), the model infers an output sequence that represents its constituent analysis.

For Spanish in particular, a key related work is the doctoral thesis of \cite{Chiruzzo2020} and the subsequent study by \cite{chiruzzo-wonsever-2020-statistical}, which compared different methods for constituency parsing. Chiruzzo used one of the most expressive representations of natural languages, the \emph{Head-Driven Phrase Structure Grammar} (HPSG) \cite{PollardSag94}, which encodes not only syntactic structures but also semantic properties. For model training, Chiruzzo used the AnCora-ES corpus developed by \cite{taule2016ancora}. Their evaluations showed that the LSTM-based approach was the most efficient.

In the case of English, significant improvement has been observed when the LSTM encoder is replaced with a transformer architecture employing self-attention, enabling the model to capture global context without recurrent networks, as explained in \cite{DBLP:journals/corr/abs-1805-01052}. For this reason, in the present work, transformer-based models are used instead of LSTMs for constituency parsing, marking the first contribution in this line for the specific case of Spanish.

Large language models available on platforms such as Hugging Face, which have greatly advanced NLP research, are developed through large-scale self-supervised pretraining, as described in \cite{DBLP:journals/corr/abs-1810-04805}. However, these large models must be fine-tuned to increase their effectiveness in specific application domains, as demonstrated in many studies, including \cite{DBLP:journals/corr/DaiL15a}, \cite{DBLP:journals/corr/abs-1802-05365}, \cite{Radford2018ImprovingLU}, and \cite{DBLP:journals/corr/abs-1801-06146}. For this reason, the present work began with the selection and fine-tuning of several large language models, as explained in the following section.

\section{Methodology}\label{sec:desarrollo}
To fine-tune a language model, a corpus suited to the target task is required. This work uses the AnCora-ES corpus as the starting point, consisting of approximately 500,000 words and 17,300 sentences, mostly composed of newspaper articles. It contains morphological, syntactic, and semantic annotations in XML format, as well as entity identification and coreference between constituents. For this study, only the XML tags and attributes identifying syntactic and, in some cases, morphological functions were used.

The corpus was adapted to perform a syntactic analysis similar to that taught in Spanish classrooms, following the notation of the \emph{Nueva gramática de la lengua española} \cite{nuevagramatica2011}. The adapted corpus was represented in a format similar to that of the Penn Treebank \cite{10.5555/972470.972475}, where syntactic structures are delimited by parentheses containing a label that identifies the structure followed by elements separated by single spaces representing its content. Parentheses were replaced with square brackets to avoid confusion with punctuation marks in Spanish. An example of the notation used is shown below:

\begin{lstlisting}[style=codigo,basicstyle=\ttfamily\footnotesize]
<s>The cup final between England and Germany had a positive effect, 
although it went almost unnoticed.</s>
<s>[Compound.Sentence [NP/S [Det The] [N final] [PP/CN [P of] [NP/T [N cup]]] 
[PP/CN [P between] [NP/T [N England and Germany]]]] [VP/PV [NP had] 
[NP/CD [Det a] [N effect] [AdjP/CN [Adj positive]]] 
[AdvP/AP [Punct ,] [conj although] [Subj it] 
[VP/PV [NP went] [AdjP/PVO [AdvP [Adv almost]] [Adj unnoticed]]]]] [Punct .]]
</s>
\end{lstlisting}

Four Hugging Face models were fine-tuned using this corpus: \texttt{bigscience/bloom-560m}, \texttt{bigscience/bloom-1b1} \cite{scao2022bloom}, \texttt{PlanTL-GOB-ES/gpt2-base-bne}, and \texttt{PlanTL-GOB-ES/gpt2-large-bne} \cite{gutierrez2021maria}. Table \ref{tab:models} shows their main characteristics. It is worth noting that a larger number of parameters does not necessarily guarantee better results. The maximum token limit of the GPT-2–based models is lower than that of the Bloom models. GPT-2 models are limited to 512 input tokens, meaning the entire corpus cannot be used, as the longest sentence contains 1,239 tokens. Therefore, the dataset used for GPT-2 models contains 15,035 sentences, while the Bloom models use the full 17,300 sentences.  

Two training and evaluation approaches were conducted: one using all sentences in the corpus, and another limited to 512-token sentences. In both cases, 80\% of the data was used for training and 20\% for testing.

\begin{figure*}
\centering
\begin{subfigure}[b]{0.4\textwidth}
\begin{tikzpicture}[scale=0.7]
\begin{semilogyaxis}[
    xlabel={Epoch},
    ylabel={Loss},
    xmin=0, xmax=5,
    ymax=6,
    legend pos=north east,
    ymajorgrids=true,
    grid style=dashed,
]

\addplot[
    color=blue,
    ]
    coordinates {
    {(0.02, 5.9938)(0.04, 1.8926)(0.06, 0.4424)(0.08, 0.3736)(0.1, 0.3429)(0.12, 0.3622)(0.14, 0.3546)(0.17, 0.3046)(0.19, 0.2997)(0.21, 0.3161)(0.23, 0.312)(0.25, 0.2937)(0.27, 0.3072)(0.29, 0.2962)(0.31, 0.2883)(0.33, 0.2993)(0.35, 0.286)(0.37, 0.3034)(0.39, 0.2945)(0.41, 0.2917)(0.43, 0.2837)(0.46, 0.2729)(0.48, 0.2914)(0.5, 0.2669)(0.52, 0.2764)(0.54, 0.2753)(0.56, 0.2612)(0.58, 0.2829)(0.6, 0.2679)(0.62, 0.2761)(0.64, 0.2789)(0.66, 0.2746)(0.68, 0.2588)(0.7, 0.2529)(0.72, 0.2693)(0.75, 0.2714)(0.77, 0.2702)(0.79, 0.2763)(0.81, 0.281)(0.83, 0.2717)(0.85, 0.2508)(0.87, 0.2666)(0.89, 0.2763)(0.91, 0.2627)(0.93, 0.2633)(0.95, 0.2686)(0.97, 0.2435)(0.99, 0.2335)(1.01, 0.2402)(1.04, 0.1977)(1.06, 0.1855)(1.08, 0.1981)(1.1, 0.1984)(1.12, 0.198)(1.14, 0.2102)(1.16, 0.2129)(1.18, 0.1974)(1.2, 0.1845)(1.22, 0.2125)(1.24, 0.1995)(1.26, 0.2006)(1.28, 0.21)(1.3, 0.2)(1.33, 0.195)(1.35, 0.1996)(1.37, 0.2013)(1.39, 0.203)(1.41, 0.2138)(1.43, 0.2045)(1.45, 0.2014)(1.47, 0.2124)(1.49, 0.1946)(1.51, 0.2114)(1.53, 0.1954)(1.55, 0.2176)(1.57, 0.2048)(1.59, 0.2103)(1.61, 0.2006)(1.64, 0.2087)(1.66, 0.2084)(1.68, 0.2192)(1.7, 0.2054)(1.72, 0.2003)(1.74, 0.2052)(1.76, 0.2067)(1.78, 0.2111)(1.8, 0.2025)(1.82, 0.2061)(1.84, 0.2031)(1.86, 0.1981)(1.88, 0.1971)(1.9, 0.2005)(1.93, 0.2054)(1.95, 0.211)(1.97, 0.2124)(1.99, 0.197)(2.01, 0.1789)(2.03, 0.1342)(2.05, 0.1394)(2.07, 0.1384)(2.09, 0.1392)(2.11, 0.1364)(2.13, 0.1322)(2.15, 0.1405)(2.17, 0.1353)(2.19, 0.1447)(2.22, 0.1355)(2.24, 0.138)(2.26, 0.1419)(2.28, 0.1394)(2.3, 0.1313)(2.32, 0.1324)(2.34, 0.1353)(2.36, 0.1409)(2.38, 0.1363)(2.4, 0.139)(2.42, 0.1391)(2.44, 0.1385)(2.46, 0.1487)(2.48, 0.1392)(2.51, 0.1385)(2.53, 0.1407)(2.55, 0.1352)(2.57, 0.1443)(2.59, 0.13)(2.61, 0.1422)(2.63, 0.1444)(2.65, 0.1441)(2.67, 0.1341)(2.69, 0.1358)(2.71, 0.1405)(2.73, 0.1401)(2.75, 0.1426)(2.77, 0.1287)(2.8, 0.1389)(2.82, 0.1307)(2.84, 0.1335)(2.86, 0.1395)(2.88, 0.1354)(2.9, 0.1418)(2.92, 0.1371)(2.94, 0.1367)(2.96, 0.1435)(2.98, 0.1398)(3.0, 0.1356)(3.02, 0.0819)(3.04, 0.0798)(3.06, 0.0854)(3.08, 0.0906)(3.11, 0.0849)(3.13, 0.0846)(3.15, 0.0839)(3.17, 0.0819)(3.19, 0.0721)(3.21, 0.0805)(3.23, 0.08)(3.25, 0.0813)(3.27, 0.0829)(3.29, 0.0819)(3.31, 0.0835)(3.33, 0.0849)(3.35, 0.0827)(3.37, 0.0891)(3.4, 0.0821)(3.42, 0.0821)(3.44, 0.08)(3.46, 0.0798)(3.48, 0.0833)(3.5, 0.0852)(3.52, 0.0701)(3.54, 0.0838)(3.56, 0.0848)(3.58, 0.0834)(3.6, 0.0782)(3.62, 0.0834)(3.64, 0.0819)(3.66, 0.0852)(3.69, 0.0819)(3.71, 0.093)(3.73, 0.081)(3.75, 0.0804)(3.77, 0.0845)(3.79, 0.0822)(3.81, 0.0841)(3.83, 0.079)(3.85, 0.0759)(3.87, 0.0863)(3.89, 0.0787)(3.91, 0.0863)(3.93, 0.0772)(3.95, 0.078)(3.98, 0.0798)(4.0, 0.0794)(4.02, 0.0546)(4.04, 0.0443)(4.06, 0.0487)(4.08, 0.049)(4.1, 0.0491)(4.12, 0.0493)(4.14, 0.0481)(4.16, 0.052)(4.18, 0.05)(4.2, 0.0472)(4.22, 0.0494)(4.24, 0.046)(4.27, 0.046)(4.29, 0.0448)(4.31, 0.0486)(4.33, 0.0495)(4.35, 0.049)(4.37, 0.0486)(4.39, 0.0501)(4.41, 0.046)(4.43, 0.0481)(4.45, 0.0472)(4.47, 0.0486)(4.49, 0.05)(4.51, 0.0457)(4.53, 0.0456)(4.55, 0.0473)(4.58, 0.0491)(4.6, 0.0506)(4.62, 0.0488)(4.64, 0.0489)(4.66, 0.0466)(4.68, 0.0446)(4.7, 0.0477)(4.72, 0.0465)(4.74, 0.0481)(4.76, 0.0466)(4.78, 0.044)(4.8, 0.0466)(4.82, 0.0445)(4.84, 0.0435)(4.87, 0.0453)(4.89, 0.0479)(4.91, 0.0444)(4.93, 0.0457)(4.95, 0.0465)(4.97, 0.0443)(4.99, 0.0472)}
    };    
    \addlegendentry{gpt2-base-bne}
    
\addplot[
    color=red,
    ]
    coordinates {
    {(0.02, 5.9869)(0.04, 0.9982)(0.06, 0.3696)(0.08, 0.3113)(0.1, 0.2925)(0.12, 0.296)(0.14, 0.2958)(0.17, 0.292)(0.19, 0.2894)(0.21, 0.3026)(0.23, 0.3032)(0.25, 0.2798)(0.27, 0.2791)(0.29, 0.2782)(0.31, 0.2677)(0.33, 0.2812)(0.35, 0.2999)(0.37, 0.2713)(0.39, 0.2984)(0.41, 0.2526)(0.43, 0.2815)(0.46, 0.2696)(0.48, 0.2736)(0.5, 0.2845)(0.52, 0.2691)(0.54, 0.2845)(0.56, 0.2685)(0.58, 0.2771)(0.6, 0.2632)(0.62, 0.2659)(0.64, 0.2638)(0.66, 0.2716)(0.68, 0.2627)(0.7, 0.2563)(0.72, 0.27)(0.75, 0.2618)(0.77, 0.2635)(0.79, 0.2553)(0.81, 0.2689)(0.83, 0.2739)(0.85, 0.2576)(0.87, 0.2634)(0.89, 0.2643)(0.91, 0.2486)(0.93, 0.2579)(0.95, 0.2507)(0.97, 0.2634)(0.99, 0.267)(1.01, 0.1828)(1.04, 0.1388)(1.06, 0.1571)(1.08, 0.1549)(1.1, 0.1415)(1.12, 0.1393)(1.14, 0.1647)(1.16, 0.1548)(1.18, 0.1552)(1.2, 0.1576)(1.22, 0.1538)(1.24, 0.1551)(1.26, 0.1524)(1.28, 0.1465)(1.3, 0.1458)(1.33, 0.1503)(1.35, 0.1574)(1.37, 0.1453)(1.39, 0.1636)(1.41, 0.1506)(1.43, 0.1715)(1.45, 0.1563)(1.47, 0.1454)(1.49, 0.1421)(1.51, 0.1461)(1.53, 0.1579)(1.55, 0.1592)(1.57, 0.1594)(1.59, 0.1577)(1.61, 0.1518)(1.64, 0.1548)(1.66, 0.1545)(1.68, 0.1493)(1.7, 0.1519)(1.72, 0.1596)(1.74, 0.1608)(1.76, 0.1486)(1.78, 0.1568)(1.8, 0.1564)(1.82, 0.156)(1.84, 0.1588)(1.86, 0.1548)(1.88, 0.1558)(1.9, 0.1511)(1.93, 0.151)(1.95, 0.1654)(1.97, 0.1671)(1.99, 0.1509)(2.01, 0.1134)(2.03, 0.0595)(2.05, 0.0612)(2.07, 0.0635)(2.09, 0.0645)(2.11, 0.0594)(2.13, 0.0639)(2.15, 0.0601)(2.17, 0.0595)(2.19, 0.0628)(2.22, 0.058)(2.24, 0.0653)(2.26, 0.062)(2.28, 0.0637)(2.3, 0.0654)(2.32, 0.0628)(2.34, 0.0668)(2.36, 0.0592)(2.38, 0.0578)(2.4, 0.0659)(2.42, 0.0622)(2.44, 0.0666)(2.46, 0.0647)(2.48, 0.0601)(2.51, 0.0639)(2.53, 0.0586)(2.55, 0.0577)(2.57, 0.0608)(2.59, 0.0667)(2.61, 0.064)(2.63, 0.0621)(2.65, 0.0612)(2.67, 0.0617)(2.69, 0.0611)(2.71, 0.0656)(2.73, 0.0633)(2.75, 0.064)(2.77, 0.063)(2.8, 0.0665)(2.82, 0.0593)(2.84, 0.0599)(2.86, 0.0657)(2.88, 0.0612)(2.9, 0.0577)(2.92, 0.0672)(2.94, 0.061)(2.96, 0.0639)(2.98, 0.0628)(3.0, 0.0602)(3.02, 0.0342)(3.04, 0.0335)(3.06, 0.0328)(3.08, 0.0336)(3.11, 0.0348)(3.13, 0.0322)(3.15, 0.0344)(3.17, 0.0341)(3.19, 0.0352)(3.21, 0.0334)(3.23, 0.0349)(3.25, 0.034)(3.27, 0.0357)(3.29, 0.0349)(3.31, 0.0342)(3.33, 0.0327)(3.35, 0.0347)(3.37, 0.034)(3.4, 0.0358)(3.42, 0.0352)(3.44, 0.0343)(3.46, 0.037)(3.48, 0.0334)(3.5, 0.0373)(3.52, 0.0342)(3.54, 0.0359)(3.56, 0.0348)(3.58, 0.035)(3.6, 0.0354)(3.62, 0.0345)(3.64, 0.035)(3.66, 0.0345)(3.69, 0.0327)(3.71, 0.0343)(3.73, 0.0341)(3.75, 0.0333)(3.77, 0.0346)(3.79, 0.0336)(3.81, 0.0358)(3.83, 0.034)(3.85, 0.0351)(3.87, 0.0333)(3.89, 0.0339)(3.91, 0.0327)(3.93, 0.0353)(3.95, 0.0326)(3.98, 0.0336)(4.0, 0.0336)(4.02, 0.0265)(4.04, 0.0261)(4.06, 0.0255)(4.08, 0.0254)(4.1, 0.0257)(4.12, 0.0249)(4.14, 0.0259)(4.16, 0.0253)(4.18, 0.0253)(4.2, 0.0251)(4.22, 0.0256)(4.24, 0.0258)(4.27, 0.026)(4.29, 0.0254)(4.31, 0.0259)(4.33, 0.0257)(4.35, 0.0249)(4.37, 0.0249)(4.39, 0.0255)(4.41, 0.0267)(4.43, 0.025)(4.45, 0.0252)(4.47, 0.0253)(4.49, 0.0251)(4.51, 0.0256)(4.53, 0.0255)(4.55, 0.0251)(4.58, 0.0258)(4.6, 0.0252)(4.62, 0.0246)(4.64, 0.0251)(4.66, 0.0252)(4.68, 0.0245)(4.7, 0.0256)(4.72, 0.0249)(4.74, 0.0245)(4.76, 0.0247)(4.78, 0.0244)(4.8, 0.0256)(4.82, 0.0243)(4.84, 0.0246)(4.87, 0.0256)(4.89, 0.0244)(4.91, 0.0236)(4.93, 0.0252)(4.95, 0.0245)(4.97, 0.0241)(4.99, 0.0253)}
    };
    \addlegendentry{gpt2-large-bne}
    
\end{semilogyaxis}
\end{tikzpicture}
\caption{gpt2 model}
\end{subfigure}
\hfill
\begin{subfigure}[b]{0.4\textwidth}
\begin{tikzpicture}[scale=0.7]
\begin{semilogyaxis}[
    xlabel={Época},
    ylabel={Error},
    xmin=0, xmax=5,
    ymax=6,
    legend pos=north east,
    ymajorgrids=true,
    grid style=dashed,
]
    
\addplot[
    color=blue,
    ]
    coordinates {
    {(0.02,2.9432)(0.04,0.2275)(0.05,0.1816)(0.07,0.1875)(0.09,0.1894)(0.11,0.1818)(0.13,0.189)(0.14,0.186)(0.16,0.1824)(0.18,0.195)(0.2,0.1796)(0.22,0.1794)(0.23,0.194)(0.25,0.1836)(0.27,0.1853)(0.29,0.1766)(0.31,0.1806)(0.33,0.1818)(0.34,0.1734)(0.36,0.1704)(0.38,0.1772)(0.4,0.181)(0.42,0.1773)(0.43,0.1675)(0.45,0.1621)(0.47,0.1666)(0.49,0.1495)(0.51,0.1566)(0.52,0.1556)(0.54,0.1624)(0.56,0.1569)(0.58,0.158)(0.6,0.1464)(0.61,0.1591)(0.63,0.1574)(0.65,0.1569)(0.67,0.1554)(0.69,0.1674)(0.7,0.1762)(0.72,0.1477)(0.74,0.1494)(0.76,0.1561)(0.78,0.1476)(0.79,0.1534)(0.81,0.1461)(0.83,0.1496)(0.85,0.1608)(0.87,0.1588)(0.89,0.1478)(0.9,0.1481)(0.92,0.1525)(0.94,0.1455)(0.96,0.1482)(0.98,0.1478)(0.99,0.1609)(1.01,0.1246)(1.03,0.1085)(1.05,0.1097)(1.07,0.104)(1.08,0.1091)(1.1,0.1097)(1.12,0.1072)(1.14,0.1065)(1.16,0.1064)(1.17,0.1154)(1.19,0.1104)(1.21,0.1041)(1.23,0.1011)(1.25,0.1114)(1.26,0.1051)(1.28,0.1058)(1.3,0.1052)(1.32,0.1112)(1.34,0.1074)(1.35,0.1091)(1.37,0.1049)(1.39,0.1047)(1.41,0.1124)(1.43,0.1125)(1.45,0.1043)(1.46,0.1087)(1.48,0.1062)(1.5,0.1155)(1.52,0.0994)(1.54,0.1109)(1.55,0.115)(1.57,0.1031)(1.59,0.1079)(1.61,0.1136)(1.63,0.1044)(1.64,0.113)(1.66,0.1098)(1.68,0.1102)(1.7,0.1011)(1.72,0.1058)(1.73,0.1064)(1.75,0.1144)(1.77,0.1081)(1.79,0.1099)(1.81,0.1087)(1.82,0.1102)(1.84,0.1063)(1.86,0.108)(1.88,0.1097)(1.9,0.1034)(1.91,0.1161)(1.93,0.1157)(1.95,0.1072)(1.97,0.1067)(1.99,0.1056)(2.01,0.0895)(2.02,0.0619)(2.04,0.053)(2.06,0.0585)(2.08,0.056)(2.1,0.0563)(2.11,0.0552)(2.13,0.0614)(2.15,0.055)(2.17,0.0561)(2.19,0.0521)(2.2,0.0612)(2.22,0.0611)(2.24,0.0582)(2.26,0.0544)(2.28,0.0588)(2.29,0.0587)(2.31,0.0582)(2.33,0.0535)(2.35,0.0573)(2.37,0.0613)(2.38,0.0552)(2.4,0.0582)(2.42,0.0575)(2.44,0.0588)(2.46,0.0612)(2.47,0.0582)(2.49,0.0558)(2.51,0.0596)(2.53,0.0559)(2.55,0.0563)(2.57,0.0592)(2.58,0.0585)(2.6,0.063)(2.62,0.0593)(2.64,0.0555)(2.66,0.0553)(2.67,0.0562)(2.69,0.0582)(2.71,0.0598)(2.73,0.0622)(2.75,0.0589)(2.76,0.0598)(2.78,0.0578)(2.8,0.0613)(2.82,0.0581)(2.84,0.0595)(2.85,0.061)(2.87,0.0574)(2.89,0.0582)(2.91,0.0585)(2.93,0.0576)(2.94,0.0569)(2.96,0.0581)(2.98,0.059)(3.0,0.0572)(3.02,0.0307)(3.03,0.0258)(3.05,0.0272)(3.07,0.0279)(3.09,0.028)(3.11,0.027)(3.12,0.0277)(3.14,0.0286)(3.16,0.028)(3.18,0.0274)(3.2,0.028)(3.22,0.0297)(3.23,0.029)(3.25,0.028)(3.27,0.0289)(3.29,0.029)(3.31,0.0303)(3.32,0.0276)(3.34,0.029)(3.36,0.0282)(3.38,0.029)(3.4,0.0282)(3.41,0.0278)(3.43,0.0293)(3.45,0.0279)(3.47,0.0297)(3.49,0.0274)(3.5,0.0287)(3.52,0.0294)(3.54,0.0275)(3.56,0.0275)(3.58,0.0285)(3.59,0.0291)(3.61,0.0277)(3.63,0.0289)(3.65,0.0277)(3.67,0.0288)(3.68,0.0302)(3.7,0.0285)(3.72,0.0281)(3.74,0.0291)(3.76,0.0279)(3.78,0.0284)(3.79,0.0287)(3.81,0.0287)(3.83,0.0274)(3.85,0.0287)(3.87,0.0287)(3.88,0.0277)(3.9,0.0271)(3.92,0.028)(3.94,0.0287)(3.96,0.0285)(3.97,0.0278)(3.99,0.0275)(4.01,0.0222)(4.03,0.0186)(4.05,0.018)(4.06,0.0186)(4.08,0.0179)(4.1,0.0182)(4.12,0.0183)(4.14,0.0181)(4.15,0.0183)(4.17,0.0184)(4.19,0.0183)(4.21,0.0185)(4.23,0.0183)(4.24,0.0186)(4.26,0.0184)(4.28,0.0184)(4.3,0.0184)(4.32,0.019)(4.34,0.018)(4.35,0.0191)(4.37,0.0182)(4.39,0.0182)(4.41,0.019)(4.43,0.0182)(4.44,0.019)(4.46,0.0181)(4.48,0.0182)(4.5,0.018)(4.52,0.0183)(4.53,0.0181)(4.55,0.0182)(4.57,0.0178)(4.59,0.0181)(4.61,0.0185)(4.62,0.0183)(4.64,0.0184)(4.66,0.0185)(4.68,0.018)(4.7,0.0184)(4.71,0.0181)(4.73,0.018)(4.75,0.0178)(4.77,0.0178)(4.79,0.018)(4.8,0.0179)(4.82,0.0181)(4.84,0.0186)(4.86,0.0181)(4.88,0.0176)(4.9,0.0182)(4.91,0.0175)(4.93,0.0175)(4.95,0.0176)(4.97,0.0177)(4.99,0.0175)}
    };
    \addlegendentry{bloom-560m}

\addplot[
    color=red,
    ]
    coordinates {
    {(0.02,3.9683)(0.04,0.2019)(0.05,0.201)(0.07,0.1873)(0.09,0.1812)(0.11,0.1725)(0.13,0.188)(0.14,0.1887)(0.16,0.1698)(0.18,0.1798)(0.2,0.1793)(0.22,0.1815)(0.23,0.1633)(0.25,0.1789)(0.27,0.1623)(0.29,0.176)(0.31,0.1757)(0.33,0.1697)(0.34,0.1779)(0.36,0.1522)(0.38,0.1682)(0.4,0.1779)(0.42,0.1702)(0.43,0.1604)(0.45,0.1836)(0.47,0.1677)(0.49,0.173)(0.51,0.167)(0.52,0.1495)(0.54,0.1764)(0.56,0.1605)(0.58,0.1654)(0.6,0.1608)(0.61,0.1506)(0.63,0.1544)(0.65,0.1542)(0.67,0.1698)(0.69,0.1679)(0.7,0.1589)(0.72,0.1583)(0.74,0.1702)(0.76,0.156)(0.78,0.1486)(0.79,0.1534)(0.81,0.1477)(0.83,0.1605)(0.85,0.1544)(0.87,0.1628)(0.89,0.1419)(0.9,0.1496)(0.92,0.1564)(0.94,0.1547)(0.96,0.1655)(0.98,0.1527)(0.99,0.1505)(1.01,0.1155)(1.03,0.0991)(1.05,0.1036)(1.07,0.0962)(1.08,0.1069)(1.1,0.0999)(1.12,0.1033)(1.14,0.113)(1.16,0.0964)(1.17,0.101)(1.19,0.1073)(1.21,0.1078)(1.23,0.1049)(1.25,0.1077)(1.26,0.1022)(1.28,0.1051)(1.3,0.1035)(1.32,0.1041)(1.34,0.1034)(1.35,0.1056)(1.37,0.0997)(1.39,0.1119)(1.41,0.105)(1.43,0.1052)(1.45,0.1093)(1.46,0.1)(1.48,0.108)(1.5,0.1123)(1.52,0.1136)(1.54,0.1063)(1.55,0.1109)(1.57,0.1103)(1.59,0.1136)(1.61,0.1055)(1.63,0.1104)(1.64,0.1135)(1.66,0.103)(1.68,0.1063)(1.7,0.1015)(1.72,0.1059)(1.73,0.1026)(1.75,0.0999)(1.77,0.1042)(1.79,0.1098)(1.81,0.1019)(1.82,0.1105)(1.84,0.1036)(1.86,0.111)(1.88,0.1019)(1.9,0.1043)(1.91,0.108)(1.93,0.1096)(1.95,0.1024)(1.97,0.1083)(1.99,0.1011)(2.01,0.0909)(2.02,0.0511)(2.04,0.0519)(2.06,0.0515)(2.08,0.0481)(2.1,0.0486)(2.11,0.0474)(2.13,0.0539)(2.15,0.0505)(2.17,0.0555)(2.19,0.0489)(2.2,0.0517)(2.22,0.0515)(2.24,0.0501)(2.26,0.0522)(2.28,0.052)(2.29,0.0527)(2.31,0.0553)(2.33,0.0517)(2.35,0.0565)(2.37,0.0536)(2.38,0.0545)(2.4,0.0496)(2.42,0.0547)(2.44,0.0532)(2.46,0.0547)(2.47,0.0539)(2.49,0.0511)(2.51,0.0481)(2.53,0.0518)(2.55,0.051)(2.57,0.0511)(2.58,0.0494)(2.6,0.055)(2.62,0.0526)(2.64,0.0518)(2.66,0.0492)(2.67,0.0518)(2.69,0.0526)(2.71,0.0556)(2.73,0.0529)(2.75,0.056)(2.76,0.0539)(2.78,0.0533)(2.8,0.054)(2.82,0.0494)(2.84,0.054)(2.85,0.0534)(2.87,0.0552)(2.89,0.0524)(2.91,0.0523)(2.93,0.0481)(2.94,0.0542)(2.96,0.0537)(2.98,0.0527)(3.0,0.0506)(3.02,0.0271)(3.03,0.0267)(3.05,0.0247)(3.07,0.0244)(3.09,0.0261)(3.11,0.0254)(3.12,0.0261)(3.14,0.025)(3.16,0.0261)(3.18,0.0255)(3.2,0.0252)(3.22,0.0256)(3.23,0.0249)(3.25,0.0253)(3.27,0.0272)(3.29,0.0263)(3.31,0.0268)(3.32,0.0256)(3.34,0.0268)(3.36,0.0269)(3.38,0.0267)(3.4,0.0258)(3.41,0.0269)(3.43,0.0261)(3.45,0.028)(3.47,0.0264)(3.49,0.0259)(3.5,0.0256)(3.52,0.0256)(3.54,0.0261)(3.56,0.0277)(3.58,0.0257)(3.59,0.0256)(3.61,0.0264)(3.63,0.0268)(3.65,0.0266)(3.67,0.0266)(3.68,0.0269)(3.7,0.0277)(3.72,0.0264)(3.74,0.0267)(3.76,0.0252)(3.78,0.0263)(3.79,0.0251)(3.81,0.0261)(3.83,0.0274)(3.85,0.0251)(3.87,0.0253)(3.88,0.0259)(3.9,0.026)(3.92,0.0254)(3.94,0.0264)(3.96,0.0262)(3.97,0.0255)(3.99,0.0243)(4.01,0.022)(4.03,0.0177)(4.05,0.0182)(4.06,0.0183)(4.08,0.0182)(4.1,0.0184)(4.12,0.0185)(4.14,0.0187)(4.15,0.0178)(4.17,0.0177)(4.19,0.018)(4.21,0.0183)(4.23,0.018)(4.24,0.0178)(4.26,0.0191)(4.28,0.0177)(4.3,0.0182)(4.32,0.0181)(4.34,0.0178)(4.35,0.0179)(4.37,0.0179)(4.39,0.018)(4.41,0.0179)(4.43,0.0176)(4.44,0.0182)(4.46,0.018)(4.48,0.0182)(4.5,0.0185)(4.52,0.018)(4.53,0.0179)(4.55,0.0183)(4.57,0.0176)(4.59,0.018)(4.61,0.0177)(4.62,0.0181)(4.64,0.0178)(4.66,0.0175)(4.68,0.0179)(4.7,0.0174)(4.71,0.0185)(4.73,0.0182)(4.75,0.0176)(4.77,0.0175)(4.79,0.018)(4.8,0.0174)(4.82,0.0178)(4.84,0.0176)(4.86,0.018)(4.88,0.0181)(4.9,0.0174)(4.91,0.0177)(4.93,0.018)(4.95,0.0175)(4.97,0.0176)(4.99,0.0177)}
    };
    \addlegendentry{bloom-1b1}
    
\end{semilogyaxis}

\end{tikzpicture}
\caption{bloom models}
\end{subfigure}
\caption{Evolution of the training loss}
\label{fig:evolutionloss}
\end{figure*}
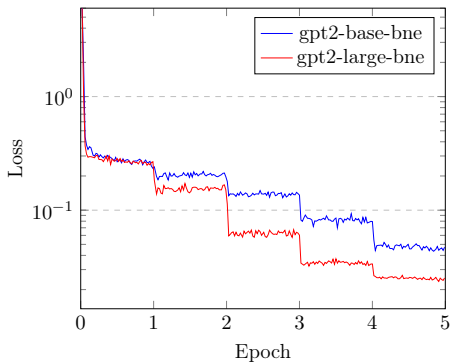
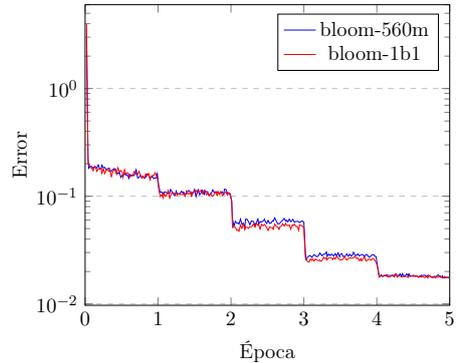

Table \ref{tab:models fine tuning} presents other fine-tuning statistics, such as training time (in seconds), memory consumption, and final loss after the last epoch. Figure \ref{fig:evolutionloss} shows the loss evolution for the four models during fine-tuning, which was conducted on a machine with an NVIDIA A100 GPU (40 GB VRAM). Five epochs were used to avoid \emph{overfitting}. However, future work will explore whether further optimization is possible without overfitting.  

\begin{figure*}[t]
\centering
\begin{subfigure}[b]{0.9\textwidth}
    \centering
    \includegraphics[width=0.9\textwidth,trim={0 0.2cm 0 0},clip]{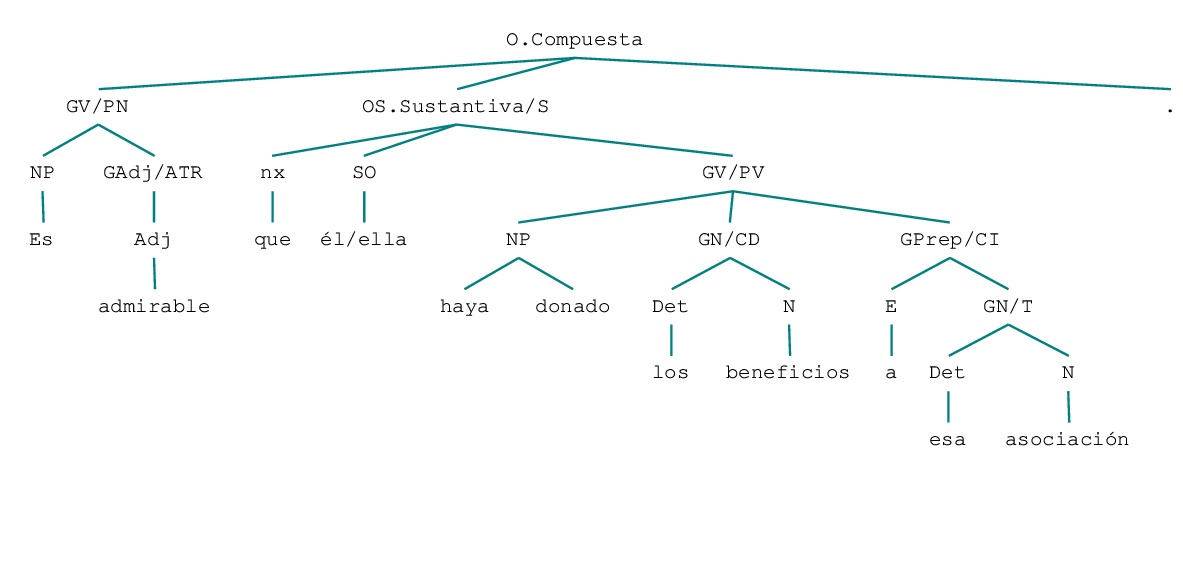}
    \caption{Compound sentence parsed by bloom-560m.}
    \label{fig:compuesta}
\end{subfigure}

\vspace{0.5cm} 

\begin{subfigure}[b]{0.9\textwidth}
    \centering
    \includegraphics[width=0.65\textwidth,trim={0 0.25cm 0 0},clip]{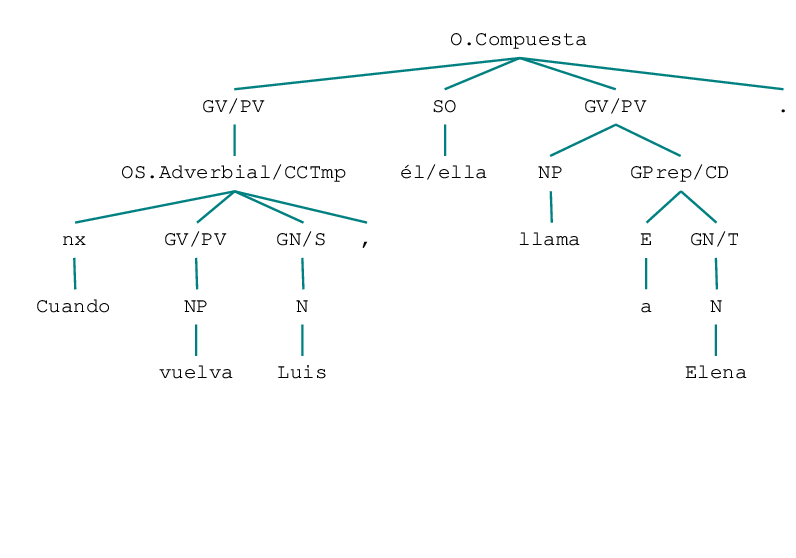}
    \caption{Ambiguous sentence parsed by gpt2-large.}
    \label{fig:ambigua}
\end{subfigure}

\caption{Examples of syntactic analyses produced by the models.}
\label{fig:ejemplos}
\end{figure*}

Table \ref{tab:models inferencia} displays the average inference time per test sentence, memory usage during inference, and the $F_1$ metric for both datasets (full and 512-token-limited). Results show minimal difference between the two. The best-performing model in terms of $F_1$ score is \texttt{gpt2-large-bne}, while \texttt{bloom-560m} achieves similar accuracy with much lower inference time. Figure \ref{fig:compuesta} illustrates a correctly parsed compound sentence, and Figure \ref{fig:ambigua} exemplifies the challenges of analyzing an ambiguous sentence (indicative vs. imperative verb).

\begin{table} 
\footnotesize{
\begin{center}
\begin{tabular}[c]{|p{2.2cm}|p{2cm}|p{2cm}|}
 \hline
 \textbf{Model} & \textbf{Parameters} & \textbf{Max Input}\\
 \hline\hline
 gpt2-base-bne & 117 million & 512 tokens\\
 \hline
 gpt2-large-bne & 774 million & 512 tokens\\
 \hline
 bloom-560m & 559 million & 2048 tokens\\
 \hline
 bloom-1b1 & 1065 million & 2048 tokens\\
 \hline
\end{tabular}
\end{center}
\caption{Characteristics of the models used}
\label{tab:models}
}
\end{table}

\begin{table} 
{\footnotesize
\begin{center}
\begin{tabular} [c]{|p{1.85cm}|p{1.5cm}|p{1cm}|p{1.4cm}|}
 \hline
\textbf{Model} & \textbf{Time} & \textbf{Loss} & \textbf{Memory}\\
 \hline\hline
    gpt2-base-bne & 1997.65 s & 0.0472 & 4295 MB\\
 \hline
    gpt2-large-bne & 11268.19 s & 0.0253 & 19883 MB\\
  \hline
    bloom-560m & 22212.93 s & 0.0175 & 23307 MB\\
 \hline
    bloom-1b1 & 36971.62 s & 0.0177 & 32867 MB\\
 \hline
\end{tabular}
\end{center}
\caption{\label{tab:models fine tuning} Fine-tuning of the models}
}
\end{table}

\begin{table}
{\scriptsize
\begin{center}
\begin{tabular}[c]{|p{1.65cm}|p{1.1cm}|p{1cm}|p{0.8cm}|p{0.8cm}|}
 \hline
\textbf{Model} & \textbf{Memory} & \textbf{Time} & \textbf{$F_1$} & \textbf{$F_1$ (512)} \\
\hline\hline
  gpt2-base-bne & 1984 MB & 1.9420 s. & 0.7234 & 0.7222\\
 \hline
  gpt2-large-bne & 4582 MB & 5.2488 s. & 0.8141 & 0.8183\\
  \hline
  bloom-560m & 3606 MB & 2.9910 s. & 0.7963 & 0.7939\\
 \hline
  bloom-1b1 & 5584 MB & 3.0467 s. & 0.7792 & 0.7665\\
 \hline
\end{tabular}
\end{center}
\caption{\label{tab:models inferencia} Inference on the AnCora-ES dataset with and without the 512-token limitation}
}
\end{table}

\section{Conclusions and Future Work}\label{sec:conclusiones}
This study aimed to explore the feasibility of constituency parsing using fine-tuned large language models under a sequence-to-sequence translation approach. The main conclusion is that this is a promising method, suggesting that large language models are capable of representing the syntactic features of the Spanish language. Future research could explore alternative methods, such as combining large language models with the CYK algorithm.

To further improve the obtained results, efforts will be made to enrich the corpus with additional sentences, particularly those corresponding to a level close to pre-university syntax learning. In addition, there are certain linguistic components recently introduced in the Spanish grammar, such as the \emph{Complemento Circunstancial de Compañía} (Circumstantial Complement of Accompaniment), which is not labeled in AnCora-ES and therefore requires new examples in the fine-tuning corpus. 

\section*{Acknowledgments}

We would like to thank the linguists Pascual Cantos, Santiago Roca, Ana Bravo, and Alejandra Valenciano for sharing their suggestions with me. Finally, I would like to express my appreciation to the MiSintaxis team members: Gonzalo Cánovas López de Molina, Laura Mateo Galindo, Tomás Bernal Beltrán, and Mario Rodríguez Béjar.

\bibliographystyle{ieeetr} 

\bibliography{biblio}

\end{document}